# A streamable large-scale clinical EEG dataset for Deep Learning

Dung Truong[1], Manisha Sinha[2], Kannan Umadevi Venkataraju[2], Michael Milham[3], Arnaud Delorme[1,4]

*Abstract*— Deep Learning has revolutionized various fields, including Computer Vision, Natural Language Processing, as well as Biomedical research. Within the field of neuroscience, specifically in electrophysiological neuroimaging, researchers are starting to explore leveraging deep learning to make predictions on their data without extensive feature engineering. The availability of large-scale datasets is a crucial aspect of allowing the experimentation of Deep Learning models. We are publishing the first large-scale clinical EEG dataset that simplifies data access and management for Deep Learning. This dataset contains eyes-closed EEG data prepared from a collection of 1,574 juvenile participants from the Healthy Brain Network. We demonstrate a use case integrating this framework, and discuss why providing such neuroinformatics infrastructure to the community is critical for future scientific discoveries.

## I. INTRODUCTION

Electroencephalography (EEG) is a neuroimaging technique that measures global-scale synchronous brain activities with high temporal resolution. The portability of EEG enables the neural underpinning of cognition in both typical laboratory settings and natural environments [1]. EEG analysis has significant implications for the understanding of human cognition, neurotechnology, as well as translational research on neuropsychiatric and neurological disorders.

Deep learning (DL) applied to large datasets is a powerful method for extracting abstract patterns from complex digital signals. DL has revolutionized fields such as Computer Vision and Natural Language Processing, and is making impacts on biomedical research [2, 3]. As a data-intensive machine learning paradigm, a particularly important aspect for the training of DL is data availability.

It is not surprising, therefore, that the rise in DL model performance in different communities comes at the point of release of large-scale datasets for that specific community [4]. We believe there is tremendous potential in applying DL directly to EEG data, and that availability of DL-ready large-scale EEG datasets for EEG can accelerate research in this field. Yet, such datasets, when available, are typically not formatted in a way that they can readily be used for DL applications. The preprocessing of such datasets often requires extensive knowledge of EEG processing, therefore limiting the pool of potential DL users.

Here we report the publication of a large-scale EEG dataset from more than a thousand subjects prepared in a format that is readily used by current DL models. This dataset is not only large in scale but also rich in metadata. We provide several labels associated with the data samples, and users may also leverage clinical information associated with this data – available upon request from the Child Mind Institute (see Methods). We set up a framework that makes the training of models straightforward, as no local data download is needed, and all data is stored on Amazon Web Service. We finally illustrate the use of this dataset in Deep Learning research via a biological sex classification task using Python and MATLAB. To the best of our knowledge, this is the first large-scale EEG dataset formatted for Deep Learning.

## II. METHODS

**Dataset.** The dataset comes from the larger data sharing project Healthy Brain Network (HBN) by the Child Mind Institute [5]. HBN is a continuing initiative focused on creating and sharing a biobank of community data from up to ten thousands of children and adolescents (ages 5-21) enriched by psychopathology assessments. Participants performed various tasks, while multimodal brain imaging (MRI, fMRI, and EEG), eye-tracking, actigraphy, voice and video data were recorded. The full dataset also includes behavioral and cognitive phenotypes, as well as genetics. We selected a subset of EEG data of 1,574 subjects during the resting state task in which participants viewed a fixation cross on the center of the screen and were instructed to open or close their eyes at various points throughout the recording period, which lasted 5 minutes. High-density EEG data were recorded in a sound-shielded room at a sampling rate of 500 Hz with a bandpass of 0.1 to 100 Hz, using a 128-channel EEG geodesic hydrogel system by Electrical Geodesics Inc. (EGI).

**Data preprocessing.** Although DL may be applied to raw EEG data without any preprocessing [2], we minimally preprocessed the data using a commonly applied data preprocessing pipeline using EEGLAB v2021 [6] functions running on MATLAB 2020b. To avoid eye saccade artifacts, we used only eye-closed data segments (~170s per subject), ignoring the first and last 3 seconds of each eye-closed period (resulting in five periods of 34 seconds). We removed the mean baseline for each data channel, down-sampled the data to 128 Hz, and subsequently band-pass filtered the data between 0.25–25 Hz (FIR filter of order 6601; 0.125 Hz and 25.125 Hz cutoff frequencies (-6 dB); zero phase, non causal). Data were re-referenced to the averaged mastoids and cleaned using the Artifact Subspace Reconstruction EEGLAB plug-in *clean_rawdata* (v2.3) [7], with the options to automatically remove artifact-dominated channels (parameters used were 5 for FlatLineCriterion, 0.7 for ChannelCriterion, and 4 for LineNoiseCriterion). Removed channels were then interpolated using 3-D spline interpolation (EEGLAB *interp.m* function). No bad portions of data were removed. Finally, we segmented eye-closed data periods into non-overlapping 2-s windows: each preprocessed 2-s epoch was used as a sample for our final dataset. Each subject provided about 81 2-s samples (mean 80.8 ± 3.32). To set up our dataset for the benchmark experiment (see Results) we sub-selected 24 channels (out of the 128 available) closest to a montage used in similar work [8]. Each sample in our dataset thus had dimension 24x256 (24 channels and 2(s) x 128(Hz) time points). No further preprocessing was performed. EEGLAB preprocessing scripts are available on the GitHub repository created for this data release (https://github.com/sccn/HBN-rest-DL).

[1] Swartz Center for Computational Neuroscience, Institute for Neural Computation, UC San Diego, La Jolla, CA, USA
[2] Appasamy Associates Private Limited, Azhiyur, Tamil Nadu, India
[3] Center for the Developing Brain, Child Mind Institute, New York, 10022, NY, USA
[4] CerCo CNRS, Paul Sabatier University, Toulouse, France

**Training, validation, and test sets**. We split the dataset into training, validation, and test sets in size ratio 60:30:10. This gave 71,300 samples (885 participants) for training, 39,868 samples (492 subjects) for validation, and 16,006 samples (197 subjects) for testing.

**Labels.** We provide all available non-clinical subject information as the labels, including sex, age, and handedness. Users can select which subject information to use as labels for their models by using the custom label selection script provided in Fig. 1 and Fig. 2. Biological sex includes binary values (0 for Male and 1 for Female), with roughly equal distribution (50%) between the two sexes in train, validation, and test sets. Summary statistics of age and handedness are provided in the respective Table 1 and Table 2 below.

|        | Train | Validation | Test |
|--------|-------|------------|------|
| Min    | 5.02  | 5.04       | 5.06 |
| Max    | 21.9  | 21.7       | 21.2 |
| Mean   | 10.5  | 10.5       | 10.7 |
| Median | 9.74  | 9.88       | 10.3 |
| Stdev  | 3.61  | 3.68       | 3.37 |

**Table 1.** Summary of age information

|        | Train | Validation | Test |
|--------|-------|------------|------|
| Min    | -100 (Fully Left-Handed) | | |
| Max    | 100 (Fully Right-Handed) | | |
| Mean   | 57.9  | 58.6       | 54.3 |
| Median | 75.6  | 75.6       | 76.7 |
| Stdev  | 50.0  | 49.4       | 56.1 |

**Table 2.** Summary of handedness information

We also provide the within-subject position of the segment in the continuous data as labels. This is particularly relevant for self-supervised learning tasks where Deep Learning models learn latent representations of the data using pretext tasks such as temporal context prediction where the models learn to predict whether two EEG segments are adjacent to each other in time [9]. Learned representations of EEG samples can then be used for downstream classification tasks and still achieve high prediction accuracy without being trained directly on the categorical labels.

The original subject IDs are also provided so that interested users can submit requests for clinical information to the child mind institute and match them with the subjects' IDs. Clinical phenotypes may be obtained through the COllaborative Informatics and Neuroimaging Suite (COINS) Data Exchange (http://coins.mrn.org/dx) or an HBN-dedicated instance of the Longitudinal Online Research and Imaging System (LORIS) located at http://data.healthybrainnetwork.org/. As the HBN phenotypic data are protected by a Data Usage Agreement (DUA), users must complete the agreement, which can be found at http://fcon_1000.projects.nitrc.org/indi/cmi_healthy_brain_network/sharing.html, and have approved by an authorized institutional official before receiving access. The intent of the HBN DUA is to ensure that data users agree to protect participant confidentiality when handling the high dimensional HBN phenotypic data and that they will agree to take the necessary measures to prevent breaches of privacy.

**Deep Learning frameworks.** We test our dataset with the Deep Learning frameworks of MATLAB and Python, two popular programming languages in the neuroscientific community. For Python, we focus on the PyTorch framework for its modular, flexible, and customizable design, conducive to experimental Deep Learning projects. Each framework comes equipped with a data handling library. Thus we designed our dataset to be easily loaded using the MATLAB *imageDatastore* [10] and PyTorch *WebDataset* [11] libraries. Using these libraries makes further data processing and augmentation straightforward. Another advantage of these libraries is the possibility of directly streaming the dataset from cloud storage, making the handling and local storing of the dataset minimal.

**Direct data streaming from cloud storage.** The dataset is hosted on Amazon Web Service (AWS) cloud resource in a public Simple Storage Service (S3) bucket (Fig. 1). Storage and access fees are covered through an agreement with AWS for freely and openly accessible scientific data. A benefit of storing data on S3 buckets is the support for online data streaming. To simplify data loading as well as to support users with limited storage and memory capacity (e.g. where the entire dataset might not fit into memory), we provide a mechanism to stream the dataset individual samples directly from the S3 bucket, using the features of MATLAB *imageDatastore* and PyTorch *WebDataset* libraries. Data users only need to provide the Unified Resource Locator (URL) of the S3 bucket and data will be streamed online during model training. Only a small portion of the data (e.g. a training batch) will be loaded into memory at a time and no local storage of the data is required. MATLAB and Python implementations to support such online data streaming are different, since their respective frameworks require different file and directory configurations, as well as different file extensions to allow speedy data parsing when streaming during model training. As such, the dataset is organized in two subdirectories *matlab* and *python* in the AWS S3 bucket. The Results section provides the specific URLs and example data loading code for the two programming languages.

For MATLAB, each sample is saved in a *.mat* file and the dataset can be streamed using the *imageDatastore* object starting from version 2021b. After creating the training, validation, and/or test set *imageDatastore* objects using their respective S3 URLs, users will need to load the corresponding label files (either from a local download or directly from S3) containing labels for all samples in the sets. We note that model training when streaming the data suffers from a significant slowdown as compared to loading the entire dataset locally. MathWorks has been notified of the issue and is working on a solution.

For the Python/PyTorch framework, the *WebDataset* library enables online data streaming. Individual samples are saved in *.npy* file format and sample-label matching is done automatically by having a separate label file for each sample, sharing the same file name as the sample file but with a different file extension *.cls*. The training, validation, and test sets are then packaged into POSIX tar archives, and only the URLs to these archives on S3 are needed to create the dataset with complete labels. Note that even though all the samples are contained in a single tarball, the *WebDataset* Python object will dynamically download individual samples as needed.

**Public computational resources.** For all computation, we used The Neuroscience Gateway (NSG). NSG provides neuroscience researchers with free high-performance computing resources of the Extreme Science and Discoveries Engineering (XSEDE) network through the San Diego Supercomputer Center [12]. NSG comes installed with popular EEG processing, Machine Learning, and Deep Learning toolboxes and libraries, including MATLAB, EEGLAB, Jupyter Notebook, PyTorch, and Tensorflow. NSG comes equipped with V100 NVidia GPU with 32 GB memory and multicore CPUs with more memory space. NSG comes with both web interface and programmatic access through RESTful APIs, with easy-to-management job submission and management scheme. Leveraging this, EEGLAB toolbox comes equipped with a plugin to submit NSG jobs directly on the EEGLAB/MATLAB environment *nsgportal* [13]. This is a public resource and we recommend users use it for DL applications to EEG data.

### III. RESULTS

Below we provide access URLs and examples for the version of the dataset prepared for Python and MATLAB. Comprehensive example scripts can be found at the GitHub repository https://github.com/sccn/HBN-rest-DL.

#### A. Python example

Python version of the dataset can be downloaded at https://childmind.s3.us-west-1.amazonaws.com/python/childmind_python.zip. The training, validation, and test set can be streamed individually via PyTorch's Webdataset library after installation using the command *pip install webdataset*. Below we provide examples of Python scripts to stream and work with the training set directly from the S3 bucket. Data users can generalize it for the validation and test set by modifying the file name accordingly.

```python
import torch
import webdataset as wds
def selectLabel(x,lbl):
    # function to select desired label
    lbl_idx = ["id","sex","age","handedness","index"].index(lbl.lower())
    x = x.decode("utf-8").split(",")[lbl_idx]
    return x if lbl_idx == 0 else float(x)

s3_url = 'https://childmind.s3.us-west-1.amazonaws.com/python/childmind_train.tar' # replace 'train' with 'val' and 'test' accordingly

train_data = wds.WebDataset(s3_url).decode().map_dict(cls=lambda x: selectLabel(x,'ID')).to_tuple("npy","cls")

# Check out first sample and its label
sample, label = next(iter(train_data))
print(f'Sample size: {sample.shape}') # (24, 256)
print(f'Label: {label}') # NDARFB908HVX
```

**Figure 1.** Example Python script to stream the dataset from AWS S3 bucket. Users select the desired label by providing the appropriate label name for the second argument of the *selectLabel* function. See GitHub repository https://github.com/sccn/HBN-rest-DL for the latest version of the code.

#### B. MATLAB example

MATLAB version of the dataset can be downloaded at https://childmind.s3.us-west-1.amazonaws.com/matlab/childmind_matlab.zip. Direct data streaming from S3 bucket is demonstrated below in Fig. 2 for the training set and can be generalized for the validation and test set accordingly.

```matlab
setenv('AWS_DEFAULT_REGION','us-west-1')

%% Custom reader function
load_sample = @(x) x.sample;
readfun = @(x) load_sample(load(x));

%% Load training labels
load('s3://childmind/matlab/train_labels.mat','-mat','train_label_info');
label_col = 3; % column index containing sex label
train_labels = categorical(cell2mat(train_label_info(:,label_col)));

%% Create datastore and assign labels
load('s3://childmind/matlab/FileSet_train.mat','-mat','fs_train');
train_imds = imageDatastore(fs_train,'FileExtensions','.mat','ReadFcn',readfun,'Labels',train_labels);

%% Preview the first sample of the datastore
sample = preview(train_imds);
fprintf('Sample size = %d, %d \n', size(sample));
fprintf('Label: Sex (1 = Female, 0 = Male) of the subject is %s \n', train_imds.Labels(1));
plot(sample');
```

**Figure 2.** Example MATLAB script to stream the dataset from AWS S3 bucket. Label is selected by providing the column index in *label_col*. See GitHub repository https://github.com/sccn/HBN-rest-DL for the latest version of the code.

**Benchmark experiment.** We ran the dataset on a model modified from VGG-16, a high-performing Computer Vision Convolutional Neural Network, for the biological sex classification task [14]. Per-subject classification accuracy was 87% (95% confidence interval 86.6% - 87.4%), achieving state-of-the-art performance. Examples of model training and evaluation through data streaming can be found in the mentioned GitHub repository.

## IV. Discussion

We provided a large EEG dataset that is Deep Learning ready. Researchers can readily download the dataset from the publicly available AWS S3 bucket. We also implemented a mechanism so that no local download is required and data can be used by direct streaming. We demonstrated the integrated framework and a biological sex classification task using the dataset. To the best of our knowledge, this is the first large-scale EEG dataset that is formatted in a Deep Learning ready way, allowing non-EEG experts to process such data.

*Caching vs streaming.* Based on the size of the data, the MATLAB or Python library may cache the data, so it can be reused across training steps. Even if this is the case, our solution offers advantages over the standard model where the data is downloaded for offline processing. The main advantage is that the data can only be partially downloaded (for example female-only samples). Another advantage is that the data can be aggregated across different datasets.

Future work will focus on further conversion of archived EEG data. The standardization of data organization and metadata annotation formats such as the Brain Imaging Data Format (BIDS) [15] and Hierarchical Event Descriptor (HED) [16] will facilitate the sharing, discoveries, and integration of large-scale datasets. We will work on automated conversion tools from such formats to DL-ready applications.

## Acknowledgments

This work was supported by NIH (R24MH120037-03 and R01NS047293-17) and by MathWorks through the MATLAB Community Toolbox Program.

Expanse supercomputer time was provided via XSEDE allocations and NSG (the Neuroscience Gateway). We thank Amitava Majumdar, Subhashini Sivagnanam, and Kenneth Yoshimoto for providing computational resources.